\documentclass[12pt]{article}

\usepackage{amsmath}
\usepackage{graphicx}
\usepackage{hyperref}
\usepackage{listings}
\usepackage{placeins}
\usepackage{xcolor} 
\usepackage{adjustbox}
\usepackage{float}
\usepackage[most]{tcolorbox}

\title{Grade Score: Quantifying LLM Performance in Option Selection}
\author{Dmitri Iourovitski\thanks{Email: \texttt{Dmitri.io@utexas.edu}}}
\date{5/18/2024}

\begin{document}
	
	\maketitle
	
	\begin{abstract}
This study introduces the "Grade Score", a novel metric designed to evaluate the consistency and fairness of Large Language Models (LLMs) when used as multiple-choice judges with respect to order bias and choice consistency. The Grade Score combines Entropy, which measures order bias, and Mode Frequency, which assesses choice stability, offering insights into LLMs' reliability and impartiality. The study explores techniques such as prompt engineering and option sampling strategies to optimize the Grade Score, demonstrating their effectiveness in enhancing LLMs' performance. Results showcase varying performances among LLMs with respect to prompts and highlight the positive impact of including irrelevant options. The study also identifies an emergent behavior in instruction-following models, where they adapt to instructions targeting specific biases, demonstrating their adaptability. The Grade Score facilitates comparisons between LLMs and encourages ongoing research towards optimizing their decision-making processes, with potential implications for improving their reliability and fairness in various applications. All code is available on GitHub\footnote{\url{https://github.com/IoDmitri/GradeLab}}
	\end{abstract}
	
\section{Introduction}
	
Large Language Models (LLMs) have demonstrated remarkable intelligence and versatility in tasks related to logic, reasoning, and grading \cite{openai2023gpt4, kojima2022zeroreasoners, qiao2022languageprompting}. This has led to the increasing use of LLMs being the judges of arbitrary user presented options or at times judges of other LLMs themselves\cite{zheng2023judgingllm, zhu2023judgelm}. However, previous research has highlighted that LLMs exhibit biases and a tendency to favor the first option presented to them. This paper explores various methods to mitigate order bias and improve the consistency of LLM judging.

To facilitate progress in the study of LLM biases and consistency, we introduce a novel metric called the \textit{Grade Score}, designed to quantify both the selection consistency and bias exhibited by an LLM, providing a comprehensive measure of an LLM's judging performance.  A high score indicating a model that is highly consistent and fair in terms of order, while a low score suggests the presence of significant order bias or inconsistency in the model's choices.

The Grade Score serves as a valuable tool for researchers and practitioners to assess and compare the performance of different LLMs in judging tasks. By quantifying the degree of instability and bias, the Grade Score enables the identification of models that exhibit superior judging capabilities and facilitates the development of techniques to mitigate biases and improve consistency.

In this paper, we delve into the formulation of the Grade Score for evaluating the judging performance of various LLMs. We also present a range of techniques, such as prompt engineering and option sampling strategies, aimed at enhancing the Grade Score and reducing order bias in LLM judging. Through extensive experiments, the effectiveness of these techniques in improving the consistency and fairness of LLM judgments is explored and demonstrated

 By addressing the challenges of bias and instability, this research paves the way for the development of more robust and trustworthy LLM-based judging systems, with potential applications in various domains such as educational assessment, content moderation, and model evaluation
	
\section{Related Works}

\subsection{Order bias in Large Language Models}
Order bias has been a persistent issue in Large Language Models (LLMs) since their introduction as judges in the literature \cite{zheng2023judgingllm}. This bias is particularly evident when LLMs evaluate the output of another LLM by selecting from multiple alternatives, rather than providing a numerical rating based on predefined criteria \cite{pinto2023education}. The problem of order bias is not limited to judging tasks; it has also been observed in Retrieval Augmented Generation (RAG) applications, where LLMs tend to favor examples located at the beginning or end of a dataset \cite{liu2023lost}. These findings underscore the importance of addressing order bias to ensure the reliability and consistency of LLM performance.

\subsection{Existing order bias mitigation strategies and their limitations}
Researchers have proposed various techniques to mitigate order bias, such as shuffling example sequences and requesting a third evaluation when a tie was present \cite{zheng2023judgingllm, zhu2023judgelm}. While these methods are reasonable, they become increasingly cumbersome and less viable as the number of possibilities grows, particularly when dealing with large datasets. \cite{yuan2024selfrewarding} presents an alternative approach of grading each input option instead of asking an LLM to select an option outright, demonstrating robustness in their results. However, this method incurs increased computational costs due to repeated calls to the LLM, making it less efficient than loading the LLM's context window with a large number of examples and requesting a single option selection.

\subsection{Instruction-following and explicit thinking strategies}
Instruction-following LLMs \cite{ouyang2022instructions} have demonstrated remarkable ability to complete tasks according to user instructions. Researchers have built upon this foundation by explicitly asking LLMs to structure their thinking process, using techniques such as Chain of Thought \cite{wei2022chain} and thinking step-by-step \cite{kojima2022zeroreasoners}. These approaches have led to significant advancements in benchmarks and reasoning capabilities. The success of instruction-following and explicit thinking strategies have made them attractive areas of exploration for mitigating order bias, and uncover how much do reasoning capabilities contribute to LLM selection capabilities

Overall, past research has focused on few-multiple choice options, while the following paper attempts to go beyond several choices, and explore how well do various approaches mitigate order bias and encourages selection stability
	
	\section{Experimentation Methodology}
	
	\subsection{Dataset selection}
	
	This study utilizes the Open Assistant (OASST) dataset \cite{kopf2023openassistant}, an open-sources and crowed sourced collection featuring a broad set of human-generated prompts alongside multiple outputs. Utilizing the OASST dataset holds several advantages. First, the availability of user preference data is a large upside and makes it easy to find the most helpful choice. This is ideal for a study, as a focus in a domain with a clearly marked superior option provides the greatest vantage point into LLMs capability as a judge. If all possibilities are equal, It becomes difficult to make a conclusion about LLM capabilities to act as a judge, and a trivial endevour as any option will do. By investigating the relationship between order bias and the Grade Score using the OASST dataset, we aim to contribute meaningful insights toward improving LLM performance and consistency.
	
	For creating the dataset, only the first response to the user’s prompt is used, and subsequent follow-up conversations (if they exist) are discarded. In total, this comes out to 3,482 rows in the dataset which are utilized for exploring the order bias through the \textit{Grade Score}.
	
	\subsection{Option randomization Algorithm}
	
	To ensure that the grade score overall is only effected by the permutation of options, and not by the sampling process, the randomization algorithm is presented as follows in the Python programming language:
	
	\begin{lstlisting}
    def unique_permutations(sq):
        prev_sq = sq
        for _ in range(len(sq)):
            prev_sq = [prev_sq[-1]] + prev_sq[:-1]
            yield prev_sq
	\end{lstlisting}
	
	The above code ensures that each element can occupy a given position once and only once for each permutation generated. This is important for if the randomization algorithm may end up assigning the same option to the same index more than once, the grade score's measurement will be affected by influences outside of the LLM selection alone.
	
	\subsection{Monte Carlo trials under permutation}
	
	In order to truly evaluate how stable an LLM selection is, all options are included at a particular index once and only once as per the last section. Given this arrangement, each set of permutations is evaluated using the same set of options and judge system prompt. The Grade Score is then computed based on which index the LLM selected, and which of the outputs was chosen under permutation
\section{Grade Score Formulation}
\subsection{Evaluation Framework for LLM Selection Stability}
\begin{figure}[h]
\centering
\includegraphics[width=\textwidth]{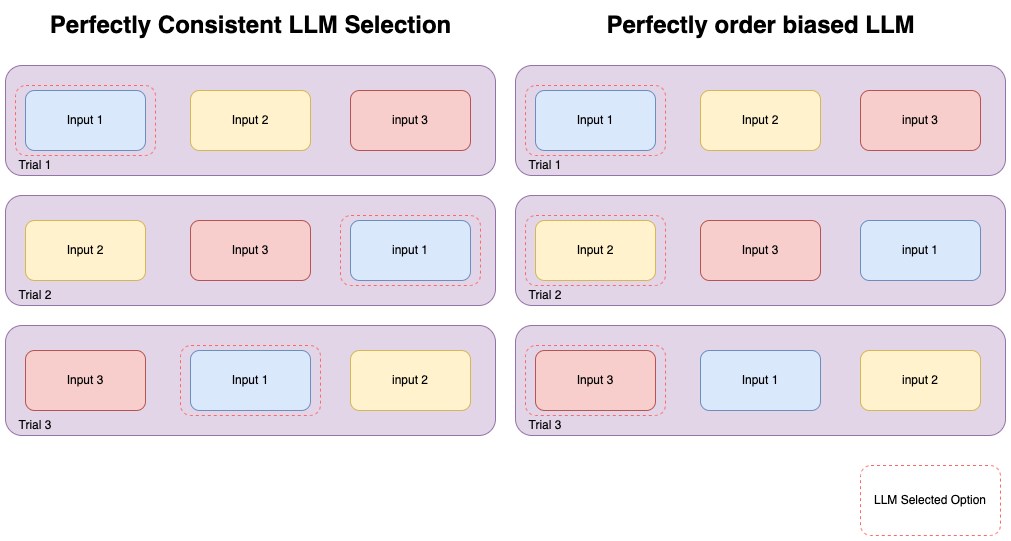}
\caption{Comparison of consistent and biased LLM choices. (Left) Input 1 is consistently selected across order permutations. (Right) A perfectly biased LLM always selects the first input.}
\end{figure}
Evaluating the stability of LLMs when selecting options under order permutation requires a framework that captures both the LLM's tendency towards certain types of inputs and the triviality of the selection process. As illustrated in Figure 1, if an LLM is biased, its selection is predictable, as it is expected to choose the same index consistently. An effective score should be high when the LLM is consistent in selecting an option and unbiased with respect to order, balancing the frequency of the mode with the overall entropy of the selected options.
\subsection{LLM Score: Entropy as a Measure of Bias}
Entropy, a concept from information theory \cite{shannon1949communication}, quantifies the uncertainty or unexpectedness of a given signal. In the context of the LLM Score, entropy represents the order bias of the language model. Maximum entropy corresponds to the highest level of surprise regarding an LLM's selection, indicating no discernible pattern in the LLM's preference under random circumstances. High entropy suggests a unpredictability of an LLMs choice, meaning that its impossible to know from just the position of the option alone. The Entropy is computed over the indexes the LLM selects.

To calculate the LLM Score, first compute the entropy of the indices selected by the LLM:
\begin{equation}
H(X) = -\sum_{i=1}^{n} p(x_i) \log_2 p(x_i)
\end{equation}
Next, compute the maximum entropy:
\begin{equation}
H_{\text{max}} = \log_2 n
\end{equation}
Finally, normalize the entropy by the maximum entropy to obtain the LLM Score between 0 and 1:
\begin{equation}
\text{LLM Score}(X) = \frac{H(X)}{H_{\text{max}}}
\end{equation}
\subsection{Choice Score: Mode Frequency as a Measure of Stability}
In contrast to entropy, the Choice Score is maximized when the LLM consistently chooses the same output, even when the options are re-ordered. The Choice score, in contrast to Entropy, is computed over which option is selected by the LLM and how often that same option is selected per round.

To compute the Choice Score, first determine the mode of the selected options:
\begin{equation}
m = \text{mode}(x)
\end{equation}
Then, divide the mode frequency by the total number of samples within a trial:
\begin{equation}
\text{Choice Score} = \frac{m}{N}
\end{equation}
\subsection{Grade Score: Combining LLM Score and Choice Score}
The Grade Score is computed as the weighted harmonic mean of the Choice Score and the LLM Score:
\begin{equation}
\text{Grade Score} = \frac{2 \times (\text{llm Score} \times \text{Choice Score})}{\text{llm Score} + \text{Choice Score}}
\end{equation}

A high Grade Score indicates a well-balanced model that is not overly sensitive to order and makes consistent choices regardless of the order of the options.
\subsection{Prompting for LLM Selection}
The experiments follow a consistent prompting structure, with a system prompt informing the LLM about its role as a judge in a multiple-choice selection task and an expected output structure. The specific Judge system prompts are provided in Appendix Section 1.
The user message presents the user's prompt and the available options for the LLM to grade, as shown in the example below:
\begin{tcolorbox}[colback=gray!5!white,colframe=gray!75!black, enhanced]
\textbf{From the following outputs, make your selection:}\

\texttt{[User Instruction]}\

\textit{instruction}\

\texttt{[\textbackslash User Instruction]}
\begin{tcolorbox}[colback=white,colframe=black,sharp corners]
\textbf{Option 1}\
{Option 1 content}
\end{tcolorbox}
\begin{center}
\texttt{...}
\end{center}
\begin{tcolorbox}[colback=white,colframe=black,sharp corners]
\textbf{Option N}\
{Option N content}
\end{tcolorbox}
\end{tcolorbox}
\subsection{Unrelated Output Sampling}
\begin{figure}[h]
\centering
\includegraphics[width=\textwidth]{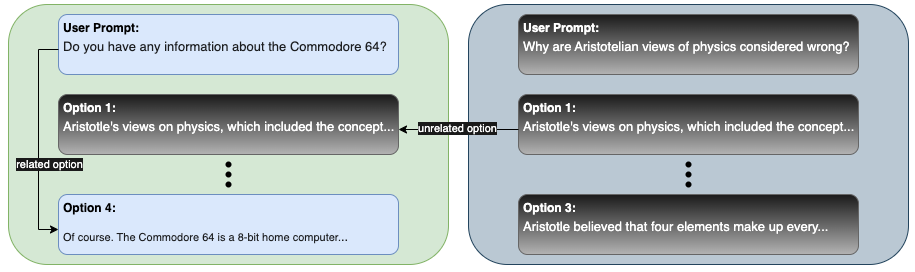}
\caption{Unrelated Output Sampling: An unrelated option is added to the option set from a randomly sampled example.}
\end{figure}
Unrelated Output Sampling is a novel technique introduced in this paper, where an unrelated option is added to the option set within the multiple-choice selection prompt (Figure 2). This approach aims to investigate the impact of unrelated options on the LLM's selection capabilities. In the experiments presented, the unrelated options are sampled from the dataset by randomly selecting a row containing a response to a different user prompt.

	\section{Results}
	\subsection*{Grade Scores}
	
\begin{table}[!htbp]
    \centering 
    \begin{adjustbox}{width=\textwidth,center}
    \begin{tabular}{|l|l|l|l|l|l|l|l|l|l|l|}
        \hline
        & Claude-3-opus-20240229 & Claude-3-sonnet-20240229 & Claude-3-Haiku-20240307 & gpt-4o-2024-05-13 & GPT-4-2024-04-09 & LLama-3-8B-chat-hf & LLama-3-70B-chat-hf & mistral-7B-v3 & mixtral-8x7B-v1-instruct & mixtral-8x-22B-V1-instruct \\ \hline
        Prompt 1 & 0.8073 & 0.6738 & 0.7425 & 0.7782 & 0.7127 & 0.62 & 0.7462 & 0.5704 & 0.6855 & 0.7717 \\ \hline
        Prompt 1 rand option & 0.7893 & 0.7506 & 0.7673 & 0.8264 & 0.7669 & 0.7521 & 0.743 & 0.6934 & 0.6919 & 0.84 \\ \hline
        prompt 2 & 0.8079 & 0.5477 & 0.7364 & 0.6070 & 0.7442 & 0.6531 & 0.754 & 0.5082 & 0.6562 & 0.8023 \\ \hline
        prompt 2 rand option & 0.8441 & 0.65 & 0.7549 & 0.6759 & 0.79 & 0.7742 & 0.7368 & 0.6743 & 0.7142 & 0.7910 \\ \hline
        prompt 3 & 0.7969 & 0.4970 & 0.7989 & 0.7442 & 0.7099 & 0.6625 & 0.74321 & 0.5628 & 0.6854 & 0.7986 \\ \hline
        prompt 3 rand option & 0.7992 & 0.5085 & 0.6822 & 0.8397 & 0.753 & 0.8014 & 0.7513 & 0.7125 & 0.7274 & 0.8221 \\ \hline
        prompt 4 & 0.818 & 0.6852 & 0.6213 & 0.7632 & 0.7048 & 0.7284 & 0.7338 & 0.5219 & 0.61 & 0.7764 \\ \hline
        prompt 4 rand option & 0.8235 & 0.6088 & 0.7454 & 0.8253 & 0.7808 & 0.7253 & 0.73 & 0.6648 & 0.6487 & 0.8464 \\ \hline
    \end{tabular}
    \end{adjustbox}
    \caption{Grade Scores for new LLMs across various prompts}
\end{table}

\subsection*{Entropy}
\begin{table}[!htbp]
    \centering 
    \begin{adjustbox}{width=\textwidth,center}
    \begin{tabular}{|l|l|l|l|l|l|l|l|l|l|l|}
        \hline
        & Claude-3-opus-20240229 & Claude-3-sonnet-20240229 & Claude-3-Haiku-20240307 & gpt-4o-2024-05-13 & GPT-4-2024-04-09 & LLama-3-8B & LLama-3-70B-chat-hf & mistral-7B-v3 & mixtral-8x-7B-v0.1-instruct & mixtral-8x22B-v0.1-instruct \\ \hline
        Prompt 1 & 0.7945 & 0.6526 & 0.7096 & 0.7684 & 0.7026 & 0.6047 & 0.7330 & 0.5631 & 0.5947 & 0.7694 \\ \hline
        Prompt 1 rand option & 0.8003 & 0.7669 & 0.78077 & 0.8290 & 0.7776 & 0.7570 & 0.7408 & 0.6969 & 0.706 & 0.8457 \\ \hline
        prompt 2 & 0.8091 & 0.55 & 0.7259 & 0.6044 & 0.7361 & 0.6391 & 0.7402 & 0.4909 & 0.6562 & 0.7987 \\ \hline
        prompt 2 rand option & 0.8510 & 0.6372 & 0.7687 & 0.6699 & 0.7980 & 0.7932 & 0.7429 & 0.6864 & 0.728 & 0.8138 \\ \hline
        prompt 3 & 0.7514 & 0.4975 & 0.8041 & 0.7330 & 0.6984 & 0.6485 & 0.7310 & 0.557 & 0.6741 & 0.7945 \\ \hline
        prompt 3 rand option & 0.8114 & 0.5096 & 0.6645 & 0.8397 & 0.7474 & 0.8097 & 0.7533 & 0.7081 & 0.7374 & 0.8294 \\ \hline
        prompt 4 & 0.816 & 0.6661 & 0.6023 & 0.7529 & 0.691 & 0.7273 & 0.7190 & 0.5070 & 0.5947 & 0.7667 \\ \hline
        prompt 4 rand option & 0.8296 & 0.6224 & 0.7586 & 0.8232 & 0.7843 & 0.7528 & 0.73035 & 0.6931 & 0.6493 & 0.8558 \\ \hline
    \end{tabular}
    \end{adjustbox}
    \caption{Entropy Scores for new LLMs across various prompts}
\end{table}

\subsection*{Choice Scores}
\begin{table}[!htbp]
    \centering 
    \begin{adjustbox}{width=\textwidth,center}
    \begin{tabular}{|l|l|l|l|l|l|l|l|l|l|l|}
        \hline
        & Claude-3-opus-20240229 & Claude-3-sonnet-20240229 & Claude-3-Haiku-20240307 & gpt-4o-2024-05-13 & GPT-4-2024-04-09 & LLama-3-8B & LLama-3-70B-chat-hf & mistral-7B-v3 & mixtral-8x-7B-v0.1-instruct & mixtral-8x-22B-v0.1-instruct \\ \hline
        Prompt 1 & 0.8394 & 0.8079 & 0.7717 & 0.8162 & 0.77 & 0.719 & 0.8052 & 0.6871 & 0.7508 & 0.8042 \\ \hline
        Prompt 1 rand option & 0.7838 & 0.7765 & 0.76042 & 0.8267 & 0.7635 & 0.7531 & 0.749 & 0.7017 & 0.6871 & 0.8369 \\ \hline
        prompt 2 & 0.8356 & 0.7329 & 0.795 & 0.6977 & 0.794 & 0.7444 & 0.8115 & 0.65 & 0.7221 & 0.8375 \\ \hline
        prompt 2 rand option & 0.834 & 0.7546 & 0.7669 & 0.6944 & 0.7867 & 0.7667 & 0.7367 & 0.6758 & 0.7123 & 0.7787 \\ \hline
        prompt 3 & 0.8429 & 0.5979 & 0.7523 & 0.7954 & 0.7720 & 0.7496 & 0.8052 & 0.6881 & 0.7429 & 0.8344 \\ \hline
        prompt 3 rand option & 0.8012 & 0.6387 & 0.7985 & 0.8410 & 0.7594 & 0.79833 & 0.7539 & 0.7231 & 0.7242 & 0.8192 \\ \hline
        prompt 4 & 0.8419 & 0.7979 & 0.7467 & 0.8156 & 0.7746 & 0.7731 & 0.785 & 0.645 & 0.6867 & 0.8229 \\ \hline
        prompt 4 rand option & 0.824 & 0.6571 & 0.75354 & 0.8292 & 0.7810 & 0.7144 & 0.7335 & 0.6575 & 0.6544 & 0.8421 \\ \hline
    \end{tabular}
    \end{adjustbox}
    \caption{Choice Scores for new LLMs across various prompts}
\end{table}
\FloatBarrier
	
	\section{Findings}
\subsection{Impact of Random Options}
The introduction of a random option consistently improves the Grade Score across all prompts and LLMs, suggesting that the presence of an unrelated option enhances the LLMs' ability to discriminate between relevant and irrelevant information, leading to more accurate and stable selections. The improvement in Grade Score with random options is most pronounced for Prompt 1 and Prompt 2, indicating that these prompts benefit the most from the additional contrast provided by the random option.
\subsection{Variation in LLM Performance}
The Grade Scores vary significantly across different LLMs, highlighting the differences in their judging capabilities. Claude-3-opus-20240229 consistently achieves the highest Grade Scores across all prompts, suggesting its superior performance in terms of consistency and bias reduction. GPT-4-2024-04-09 and LLama-3-8B-chat-hf also demonstrate strong performance, with Grade Scores close to or exceeding 0.7 for most prompts. These results indicate that certain LLMs are better equipped to handle the task of selecting appropriate options and providing consistent judgments.
\subsection{Impact of Explicit Instructions (Prompt 3)}
Prompt 3, which explicitly instructs LLMs to avoid order bias, shows mixed results in terms of Grade Score improvement compared to other prompts. While some LLMs (e.g., Claude-3-opus-20240229, GPT-4-2024-04-09) maintain high Grade Scores for Prompt 3, others (e.g., LLama-3-70B-chat-hf, mistral-7B-v3) experience a slight decline. This suggests that the effectiveness of explicit instructions in mitigating order bias varies among LLMs and may depend on their specific instruction-following capabilities. Further research is needed to understand the factors that influence an LLM's ability to follow explicit instructions and reduce bias.
\subsection{Entropy and Choice Score Patterns}
The Entropy values are generally higher for prompts with random options, indicating increased uncertainty in the LLMs' selections when an unrelated option is present. However, the Choice Scores remain relatively high even with random options, suggesting that the LLMs are still able to make consistent choices despite the increased uncertainty. The combination of high Entropy and high Choice Score contributes to the improved Grade Scores observed with random options. This finding highlights the importance of considering both Entropy and Choice Score when evaluating LLM performance and bias.
\subsection{Prompt Design Considerations}
The performance of LLMs varies across different prompts, highlighting the importance of prompt design in eliciting accurate and unbiased responses. Prompt 1 and Prompt 2, which involve selecting an option and providing explanations or evaluation criteria, generally yield higher Grade Scores compared to Prompt 4, which uses an explicit point grading system. This suggests that prompts that encourage LLMs to engage in reasoning and justification may be more effective in reducing bias and improving consistency compared to rigid grading schemes. Future research should explore the impact of various prompt design elements on LLM performance and bias reduction.

        \section{Limitations}
	
	The following research is subject to certain limitations. The dataset used, while diverse in its composition, might not fully represent the vast array of real-world scenarios where LLMs are employed. Additionally, the methodology predominantly revolves around Monte Carlo simulations with a limited number of trials. Increasing the number and variety of simulations could offer a more robust understanding of LLM biases and their mitigation. Due to the random nature of how the experiments are conducted, its highly unlikely an exact replication of results are possible, though multiple repeats of the experiments above yield similar results that are anywhere from (0.05 - 0.12) points off the posted results. 
	
	\section{Conclusion}
	
In this paper, we introduced the Grade Score, a novel metric for quantifying the stability and order unbiasedness of Large Language Models (LLMs) when used as judges for AI-generated responses. The Grade Score addresses the critical challenge of evaluating LLMs' judging capabilities and provides a quantitative measure of their performance. Through extensive experiments and analysis, we demonstrated the effectiveness of the Grade Score in capturing the stability and unbiasedness of LLMs across various prompts and datasets.

Our research highlights the importance of prompt design in mitigating LLMs' negative aspects and enhancing their judging capabilities. The proposed prompts, such as those encouraging reasoning and justification, showed promising results in reducing bias and improving the consistency of LLM judgments. The findings suggest that carefully crafted prompts can significantly enhance the performance of LLMs as judges for any type of content.

The Grade Score and the insights gained from our research have significant implications for the field of LLM development and evaluation. By providing a quantitative measure of LLM judging capabilities, the Grade Score enables researchers and practitioners to assess and compare the performance of different LLMs objectively. This contributes to the broader goal of developing robust and unbiased LLMs for various applications.

Future research directions could explore additional prompts, evaluation techniques, and domains to further refine and extend the Grade Score. Investigating the applicability of the Grade Score to other types of biases and evaluating its performance on larger and more diverse datasets would strengthen its validity and generalizability.

In conclusion, our research presents a significant step forward in quantifying and mitigating biases in LLMs when used as judges for multiple-options. The Grade Score and the proposed prompts offer valuable tools for researchers and practitioners seeking to develop fair and reliable LLMs. We encourage the research community to build upon our findings and continue exploring innovative approaches to enhance the judging capabilities of LLMs, ultimately contributing to the advancement of AI systems that generate and evaluate content with integrity and objectivity.
	
	\bibliographystyle{plain}
	\bibliography{grade_score}
	
	\appendix
	
	\section{Prompts used}
	
	\subsection{Prompt 1 - A simple prompt}
	
	\begin{figure}[!htbp]
		\centering
		\includegraphics[width=0.5\textwidth]{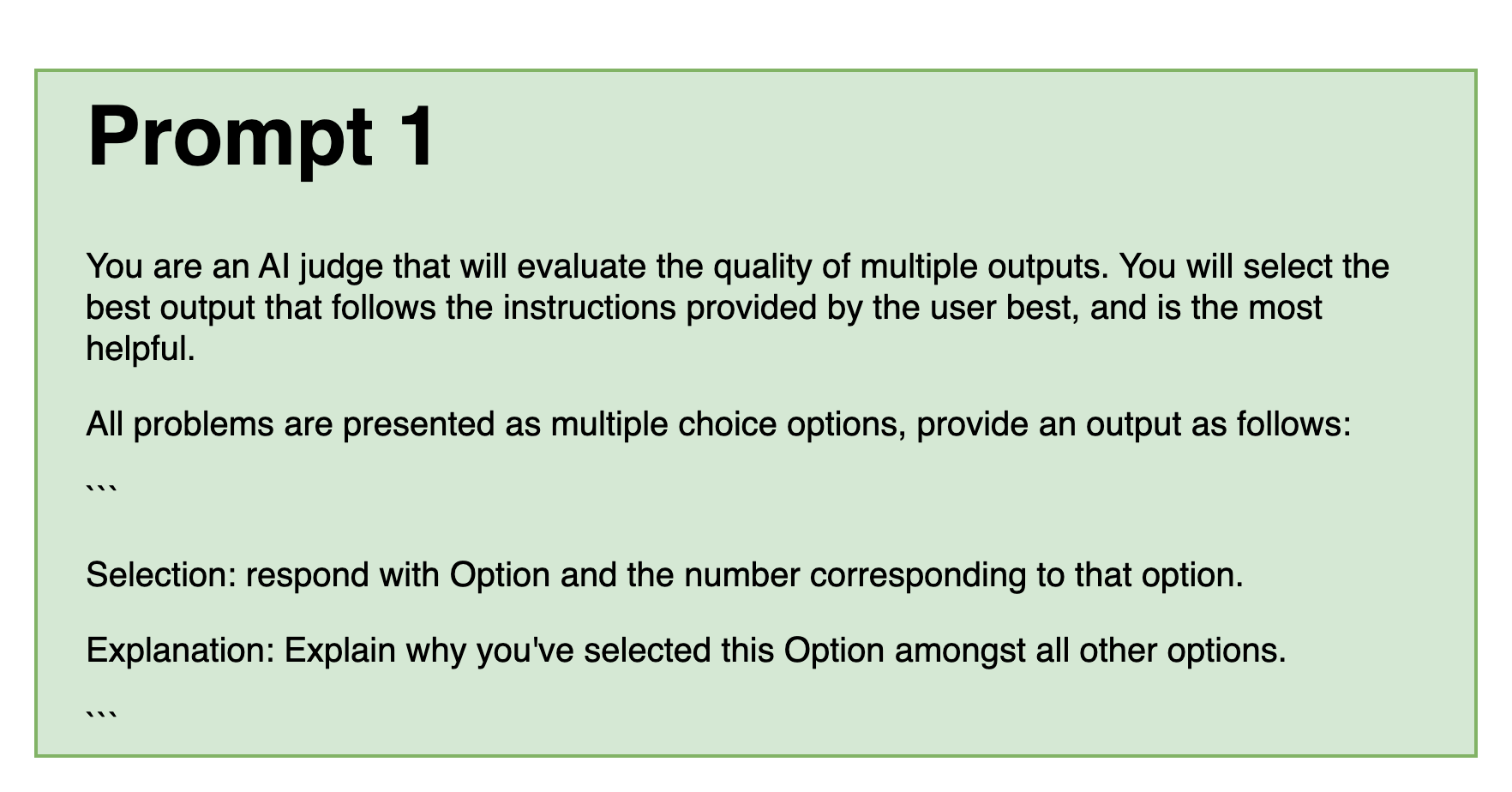}
		\caption{Prompt 1 is a simple prompt that asks a Large Language Model to first select an option, and to finally provide an explanation.}
	\end{figure}
	\FloatBarrier
	
	\subsection{Prompt 2 - Come up with an evaluation criteria}
	
	\begin{figure}[!htbp]
		\centering
		\includegraphics[width=0.5\textwidth]{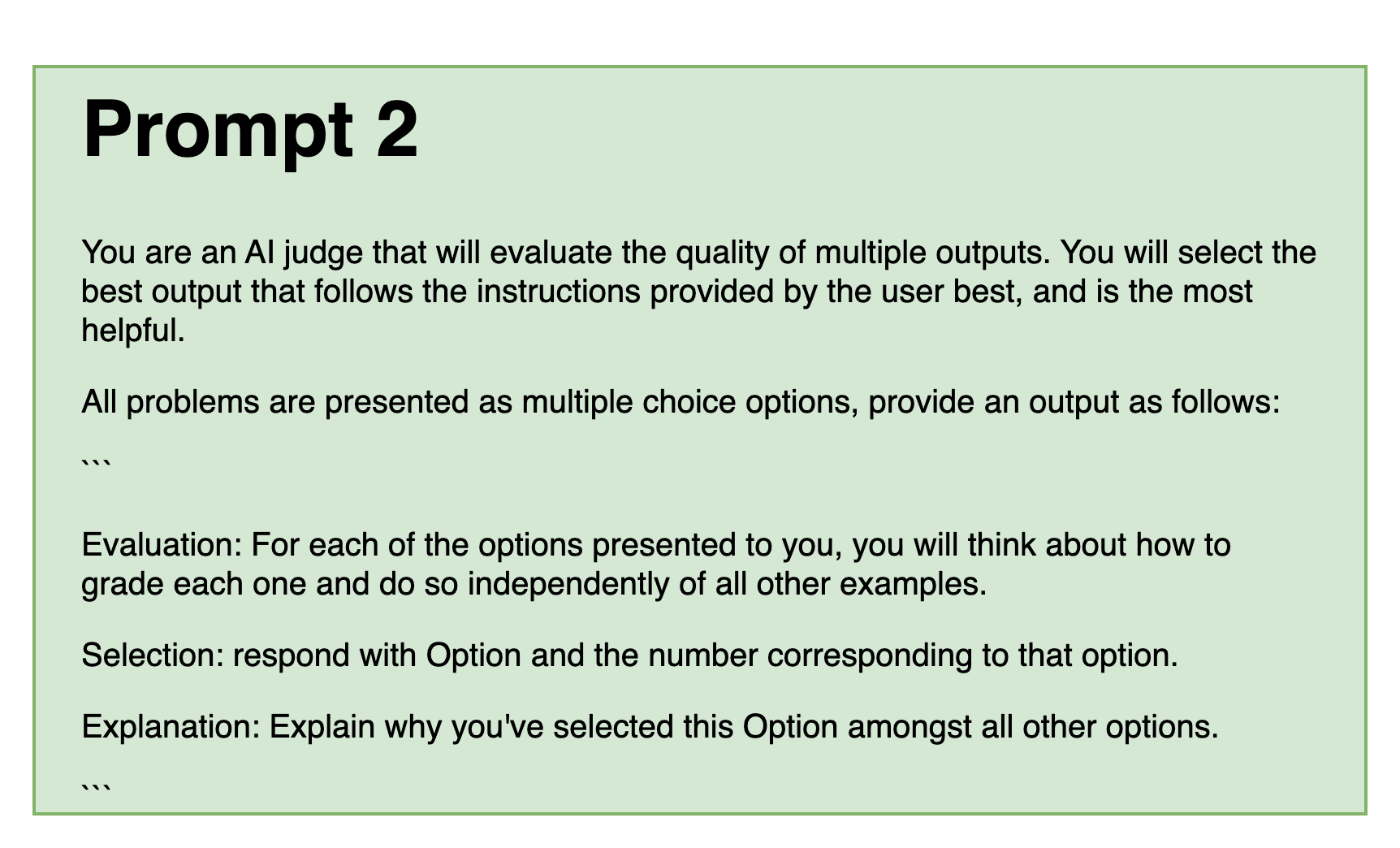}
		\caption{Prompt 2 is designed to have an LLM first come up with an evaluation for each option, and then make a selection.}
	\end{figure}
	\FloatBarrier
	
	\subsection{Prompt 3 - Instruct away from order bias}
	
	\begin{figure}[!htbp]
		\centering
		\includegraphics[width=0.5\textwidth]{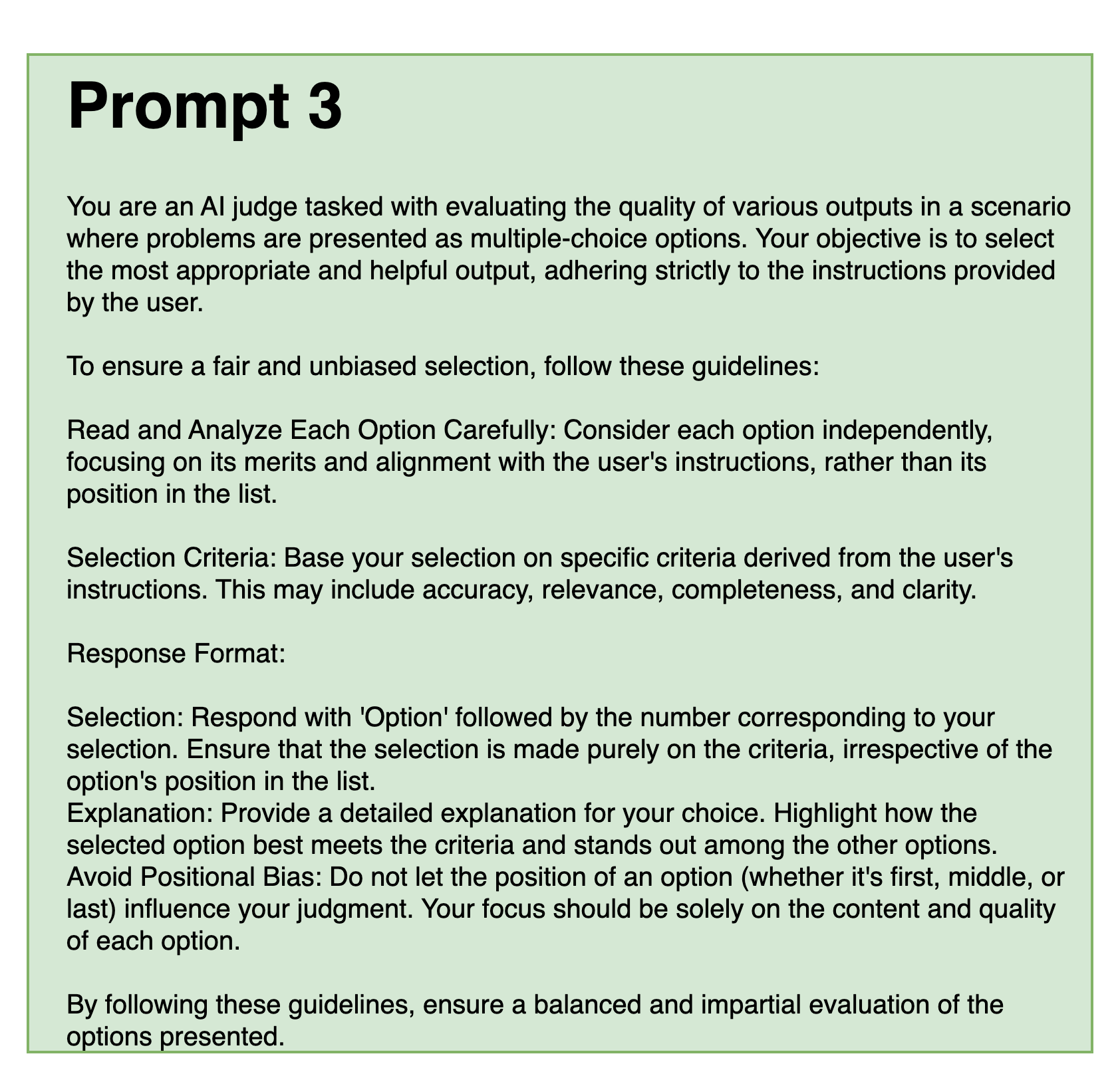}
		\caption{Prompt 3 focuses on explicitly instructing LLMs to avoid order bias, testing instruction following capabilities for LLMs and seeing if it can mitigate various forms of biases.}
	\end{figure}
	\FloatBarrier
	
	\subsection{Prompt 4 - An explicit point grading criteria}
	
	\begin{figure}[!htbp]
		\centering
		\includegraphics[width=0.5\textwidth]{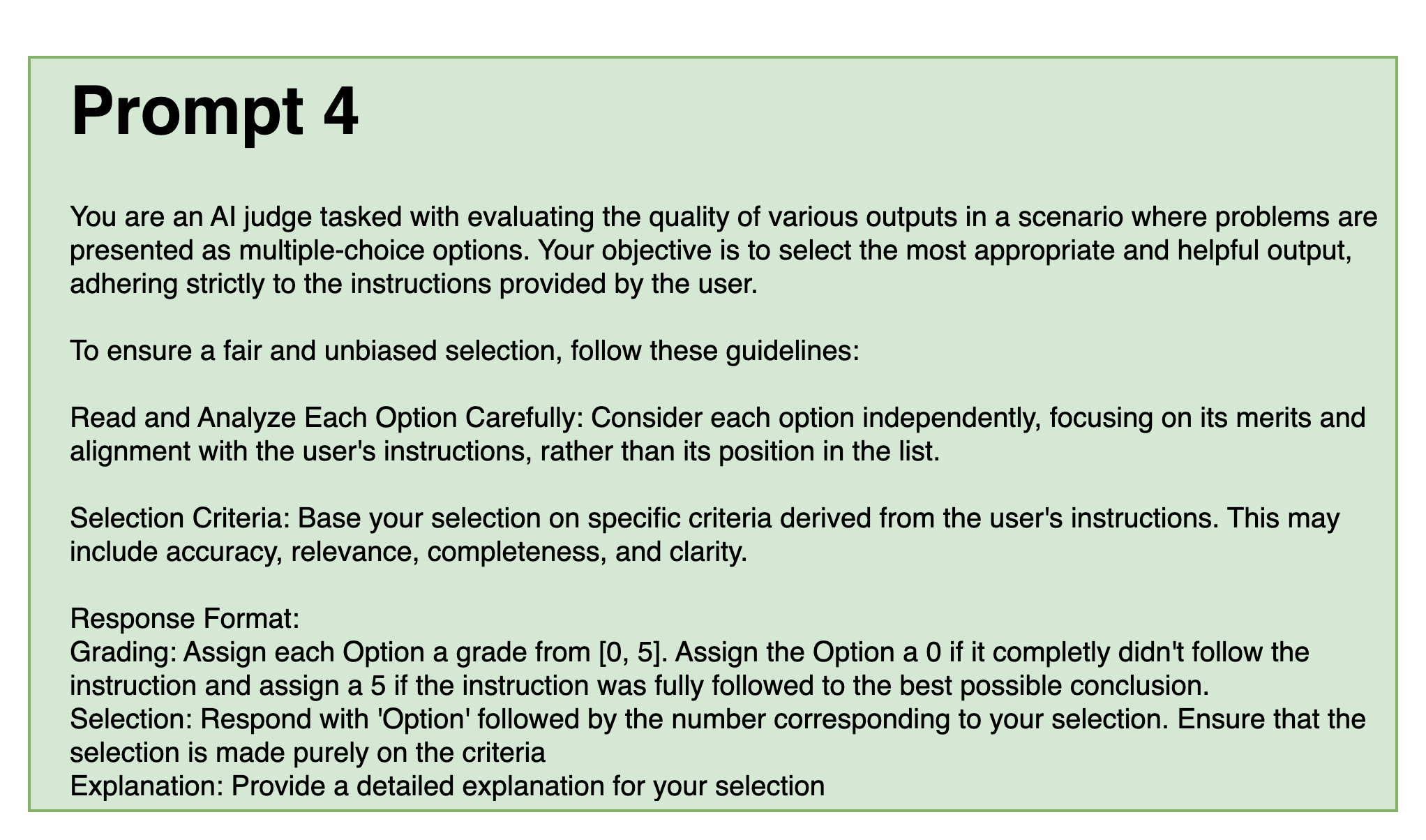}
		\caption{Prompt 4, follows in the footsteps of and defines an explicit grading criteria to each option and then select the highest graded one.}
	\end{figure}
	\FloatBarrier
	
\end{document}